# A Storage-Efficient Feature for 3D Concrete Defect Segmentation to Replace Normal Vector

Linxin Hua Ph.D.,[1] Jianghua Deng Ph.D.[2], Ye Lu Ph.D.[3]

[1] Department of Civil and Environmental Engineering, Monash University, Melbourne, VIC, Australia, 3800

[2] School of Civil Engineering & Architecture, Changzhou Institute of Technology, Changzhou, Jiangsu, China, 213032

[3] Department of Civil and Environmental Engineering, Monash University, Melbourne, VIC, Australia, 3800; email: ye.lu@monash.edu (Corresponding author)

**Abstract**

Point cloud reconstruction of damage offers an effective solution to image-based methods vulnerable to background noise, yet its application is constrained by the high volume of 3D data. This study proposes a new feature, relative angle, computed as the angle between the normal vector of a point and the average normal vector of its parent point cloud. This single-dimensional feature provides directionality information equivalent to normal vectors for concrete surface defect characteristics. Through entropy-based feature evaluation, this study demonstrates the ability of relative angle to filter out redundant information in undamaged sections while retaining effective information in damaged sections. By training and testing with PointNet++, models based on the relative angles achieved similar performance to that of models based on normal vectors while delivering 27.6% storage reduction and 83% input channel compression. This novel feature has the potential to enable larger-batch execution on resource-constrained hardware without the necessity of architectural modifications to models.

## 1 Introduction

Concrete structures inevitably experience certain degrees of damage during long-term use. Timely detection and maintenance are necessary to prevent further deterioration that could affect the serviceability of structures. Additionally, when concrete structures fail and lose their load-bearing capacity, observations of surface damage can help assess the failure and guide subsequent maintenance (Gharehbaghi et al. 2022). Currently, as numerous concrete structures have undergone extended periods of service, the demand for structural health monitoring and maintenance is gradually increasing and occupying an important position in practice (Scope et al. 2021; Van Breugel 2017).

However, in current structural health monitoring, manual inspection still dominates. This method requires inspectors to manually observe and document defects. It makes manual inspection highly demanding on human resources and carries certain risks in many inspection scenarios (Purohit et al. 2018; Spencer et al. 2019). For example, inspecting structures such as bridges involves complex site conditions and safety hazards, necessitating more elaborate and reliable processes to ensure safe inspection (Lin et al. 2021). It has created a demand for automated defect identification, detection, and recording, which is supported by rapidly developing object recognition methods and hardware (Chow et al. 2021). Most current applications rely on image-based damage identification (Huang and Liu 2024; Islam et al. 2025; Ma et al. 2024) and address various practical challenges (Alzuhiri et al. 2023; Mishra and Lourenço 2024). However, noise remains a significant issue. For example, Chu et al. (2022) reported misidentifications of water stains and paint peeling where errors come from similarities in colour and shape. Lattanzi and Miller (2014) highlighted difficulties in detecting damage under complex lighting conditions. To tackle noise challenges, Deng et al. (2020, 2025) developed optimised models that differentiate between handwriting and physical cracks. Teng et al. (2024) proposed image enhancement through cross domain translation approaches for low-resolution images collected from challenged scenarios. Although these studies have made progress in reducing noise impacts in two-dimensional images, practical requirements still demand more robust methods for reliable damage identification (Guo et al. 2024).

Using three-dimensional data for defect detection offers a solution to effectively mitigate these noise-related challenges . For instance, even in noisy settings, Hua et al. (2022, 2023) extracted defect dimensions from 3D reconstructions. In practical scenarios, the usage of 3D information enables reliable performance in identifying defects (Reyno et al. 2018; Shim et al. 2023a). Bahreini and Hammad (2024) segmented cracks and spalling in a bridge reconstruction, and Shim et al. (2023b) identified tunnel damage under low-light conditions. By leveraging differences in 3D shape and depth between damaged and undamaged areas, these approaches effectively overcome the noise challenges of 2D imaging. Many of these studies use point clouds and focus on the usage and analysis of position coordinates. In addition to this, some of them have reported a significant improvement in defect identification capability when using normal vectors of points (Loverdos and Sarhosis 2024; Wang et al. 2024). As a feature, normal vectors are considered to describe the morphology of a point and its surrounding points. This characteristic may have positive impacts on defect identification that focuses on local geometric morphology, which has been reported in many practical applications (Rani et al. 2024).

However, when using normal vectors, the volume of data for both model training and prediction increases. Normal vector covers rich local geometric information, which, although beneficial for capturing surface morphology, adds to the dimensionality of data and consequently demands higher storage and computational resources. This is especially challenging in on-site automatic structural defect detection, where device miniaturisation and real-time processing are critical (Capra et al. 2020). Several studies have explored strategies to address these challenges. Han et al. (2015) tackled this issue by leveraging normal vector differences to detect and preserve edge details during point cloud simplification, emphasising the importance of retaining critical geometric features. Hu et al. (2024) proposed a LiDAR point cloud simplification algorithm that uses a fuzzy encoding-decoding mechanism to capture uncertainty and simplify data while preserving key structural elements, though it still relies on multiple high-dimensional features. Qi et al. (2019) presented a graph-based method that enforces both feature preservation and uniformity control, yet their reliance on full normal vector data contributes to significant computational costs. Potamias et al. (2022) introduced a learnable feature-preserving approach that balances data reduction while retaining salient features via neural networks. Although these methods effectively reduce the overall data volume, the reduction is primarily attributed to a decrease in the total number of data points. Thus, they inevitably lead to the loss of information and may potentially affect the fidelity of the original geometric details.

To downsize data usage in concrete defect detection applications, this study proposes a new feature called 'relative angle'. This feature simplifies the 3D normal vectors into angle values with only one channel while preserving the full geometric integrity of the data. It is developed based on the analysis of the geometric shape of concrete damage. By keeping the essential shape and details of the targets and effectively utilising the unique characteristics of concrete damage, this feature enables more efficient and accurate damage detection without any loss of information.

The first part of this paper introduces the derivation of relative angles from position coordinates and normal vectors. Then, it describes an entropy-based feature evaluation method to compare different features of point clouds. A dataset with unfolded point clouds of 22 concrete cylinders with surface defects was developed and described in detail, covering the data collection and data preprocessing. A comparative entropy-based feature evaluation of the proposed feature against position coordinates and normal vector is then presented. After that, different feature combinations are used to train PointNet++, which is a frequently used point cloud-based neural network. The training results under different feature combinations are compared, and the improvement of the proposed feature on model performance and its potential to replace normal vectors in concrete defect identification applications are discussed. In the end, the contributions and limitations of this study are discussed, and potential future work is pointed out to further explore the possibilities of relative angle in the practice of structural health monitoring.

## 2 Relative angle as a feature

The proposed feature, relative angle, refers to the inclined angle between the normal vector of a point and the average normal vector of all the points in the same point cloud. This feature is only determined by the three-dimensional positions of points, which are applicable in general 3D data collection and extraction. It is designed based on the significant geometric difference between the damaged and undamaged sections of the concrete surface. This section illustrates the calculations of relative angles and other related characteristics of points.

### 2.1 Point coordinates

For any point cloud analysis and applications, the coordinates of points are critical features as they represent the positions of points in three-dimensional space, from which, global and local geometric shape information of the structures might be extracted. Defect identification applications usually focus on localised defects from collected damage information (Kashif Ur Rehman et al. 2016; Ndambi et al. 2002).

The 3D reconstruction of those localised concrete defects presents unique characteristics distinct from general 3D reconstruction objectives such as human body reconstruction (Correia and Brito 2023) or geological mapping (Wu et al. 2019). These characteristics arise from the inherent geometric properties of the structure and the nature of damage development. Typically, most undamaged sections of concrete structures consist of predominantly flat surfaces (Tang et al. 2011). In contrast, damaged sections exhibit a rough, uneven, and fractured surface but without significant depth changes. These characteristics are the basis for identifying damaged and undamaged regions and can be effectively utilised in applications of structural damage identification. A common practice is that researchers often align the collected concrete surface damage data to a hypothetical plane (Bajaj et al. 2022, 2024; Smith et al. 2024). For instance, in a Cartesian coordinate system, a point cloud of concrete surface damage might be aligned to the x-z plane, with the y-values representing the depth of surface damage. With alignment, the defects can be easily identified based on differences in y-axis values (Bahreini and Hammad 2024). However, the alignment operation involves the identification of undamaged surfaces and the alignment of the point cloud data. It is a process that may require additional data preprocessing and is not always straightforward to align to a single plane. For example, damage occurring at the corners of a structure may not be aligned to a plane. Since this study focuses on the damage itself and the geometric differences between the damaged regions and their surroundings, no alignment preprocessing of the point cloud data is performed (Zheng et al. 2023).

## 2.2 Point cloud normalisation

To meet the input requirements of the point cloud recognition and segmentation models, the complete point cloud is typically partitioned into smaller, fixed-size segments, which are then normalised accordingly. This step ensures that the point cloud can be segmented into appropriately sized portions as model inputs and that numerical differences between various samples and regions remain within acceptable limits.

The division of point clouds is usually carried out based on the input dimensions of the models and the objectives of the users. In contrast, point cloud normalisation can be conducted in two primary ways: global and axis-specific normalisation (Yang et al. 2020). Fig. 1 illustrates the difference between the two normalisation methods. For simplicity, a 2D view of a point cloud with 3 points is applied.

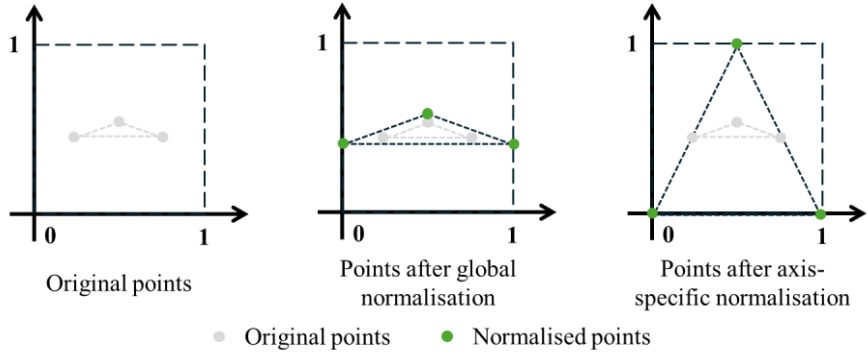

Fig. 1 Changes of points after global and axis-specific normalisation

As can be seen, global normalisation applies a uniform scale factor to all coordinates across the entire point cloud, preserving the overall shape while scaling the point cloud to reach the normalisation boundaries. In contrast, axis-specific normalisation employs separate scale factors for each coordinate axis, calculated based on the respective value ranges of axes. It rescales the point cloud along all axes to reach the normalisation boundaries. This method ensures the full range of normalised values is used for each axis. It may benefit certain segmentation tasks as it can emphasise differences along axes where there is more variation (Zheng et al. 2023). However, global normalisation maintains the geometry and proportions of the damage which may be critical for defect identification.

In this study, both normalisation methods are applied to the point cloud, and the centres of the point clouds are moved to the coordinate origin. This results in the normalised point cloud having a coordinate range of [-0.5, 0.5]. The calculation process is shown as follows.

Assume there is a point cloud $P$ that has been translated so that its centre point becomes the coordinate origin. In this case, the ranges of x, y, and z-axis of each point $p_i$, $(x_i, y_i, z_i)$, in this point cloud are $[x_{min}, x_{max}]$, $[y_{min}, y_{max}]$, and $[z_{min}, z_{max}]$ respectively. For global normalisation, the scale factor, $k_g$, is calculated as the maximum value among the ranges of each axis:

$$k_g = \max(x_{max} - x_{min}, y_{max} - y_{min}, z_{max} - z_{min}) \tag{1}$$

Thus, the coordinates of the normalised $p_i$ become:

$$x_i^g = \frac{x_i - x_{min}}{k_g} \tag{2}$$

$$y_i^g = \frac{y_i - y_{min}}{k_g} \tag{3}$$

$$z_i^g = \frac{z_i - z_{min}}{k_g} \tag{4}$$

For axis-specific normalisation, the scale factors, $k_x$, $k_y$, and $k_z$, are calculated separately according to the ranges of coordinates:

$$k_x = x_{max} - x_{min} \tag{5}$$

$$k_y = y_{max} - y_{min} \tag{6}$$

$$k_z = z_{max} - z_{min} \tag{7}$$

Thus, the coordinates of the normalised $p_i$ become:

$$x_i^a = \frac{x_i - x_{min}}{k_x} \tag{8}$$

$$y_i^a = \frac{y_i - y_{min}}{k_y} \tag{9}$$

$$z_i^a = \frac{z_i - z_{min}}{k_z} \tag{10}$$

## 2.3 Normal vector

Extensive research and practical applications have shown that normal vectors of points, as additional features beyond position coordinates, can effectively enhance the performance of neural network models in tasks such as point cloud-based identification and segmentation (Demarsin et al. 2006; Weber et al. 2010). It also works in concrete defect detection applications, effectively improving the model performance in accuracy (Chen and Cho 2022).

In this study, to achieve normal vectors, a K-Dimensional (KD) tree (Friedman et al. 1977) will first be built for the target point cloud to obtain the positional relationships among points in the unordered point cloud data (Hao et al. 2018). Based on it, $k-1$ nearest neighbours, $\boldsymbol{p_1}, \boldsymbol{p_2}, \ldots, \boldsymbol{p_{k-1}}$, can be found for a particular point, $\boldsymbol{p_0}$. The average position coordinates for those points are then calculated as follows:

$$\boldsymbol{c} = \frac{1}{k} \sum_{i=0}^{k-1} \boldsymbol{p_i} \tag{11}$$

Thus, the $3 \times 3$ covariance matrix, $\boldsymbol{C}$, is calculated:

$$\boldsymbol{C} = \frac{1}{k} \sum_{i=0}^{k-1} \left( (\boldsymbol{p_i} - \boldsymbol{c})(\boldsymbol{p_i} - \boldsymbol{c})^T \right) \tag{12}$$

For this covariance matrix, eigenvalues ($\lambda_1, \lambda_2, \lambda_3$) and corresponding eigenvectors ($\boldsymbol{v_1}, \boldsymbol{v_2}, \boldsymbol{v_3}$) are calculated. Then, the estimated normal vector of $\boldsymbol{p_0}$, $\boldsymbol{n}$, is the eigenvector corresponding to the smallest eigenvalue among $\lambda_1, \lambda_2,$ and $\lambda_3$.

## 2.4 Relative angle

Normal vector, as a commonly used feature in point cloud-based models, is capable of improving model performance through effective representation of point directions. It allows the target areas to be effectively distinguished. In concrete damage inspection practice, point cloud data is generally divided into undamaged and damaged sections. The undamaged sections are usually smoother and, in most cases, can be approximated as a plane. Moreover, the normal vectors of points in undamaged sections generally point in very similar directions. The damaged sections, on the other hand, due to the randomness of the damage development and the non-uniformity of the material, typically form irregular, rough, and uneven surfaces. This results in the normal vectors in the damaged sections showing a high degree of randomness in directions. It indicates a significant contrast to the highly consistent normal vectors of points within undamaged sections. Fig. 2 shows typical cross-sections of undamaged and damaged sections, as well as the normal vectors of some points, illustrating the directional differences of normal vectors between them.

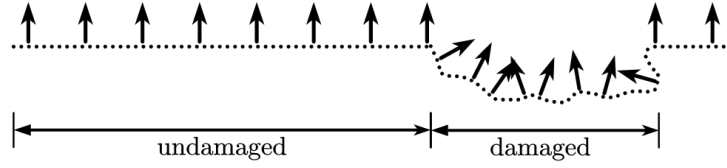

Fig. 2 Normal vectors on cross-section containing undamaged and damaged sections

It is noted that the directions of normal vectors may have the potential to distinguish between damaged and undamaged sections effectively. To more efficiently utilise this feature, this study proposes a feature designed based on the directions of normal vectors, i.e., the relative angle, which refers to the angle between the normal vector of that point and the average normal vector of all points in the same point cloud. Through this transformation, the direction of the normal vector is downsized from numerical values in three directions to a single angle value. The detailed calculations are as follows:

In a point cloud, each point $p_i$ has a corresponding normal vector $\boldsymbol{n}_i$. The average normal vector $\boldsymbol{n}_{avg}$ of the point cloud can be calculated as follows:

$$\boldsymbol{n}_{avg} = \frac{1}{N}\sum_{i=1}^{N}\boldsymbol{n}_i \tag{13}$$

, where $N$ refers to the total number of points in the point cloud. It is assumed that all the normal vectors $\boldsymbol{n}_i$ are normalised.

The relative angle, $\theta_i$, between the normal vector of each point, $\boldsymbol{n}_i$, and the average normal vector, $\boldsymbol{n}_{avg}$, of the point cloud can be calculated as follows:

$$\theta_i = \cos^{-1}(|\boldsymbol{n}_i \cdot \boldsymbol{n}_{avg}|) \tag{14}$$

## 2.5  Entropy-based feature evaluation

The proposed feature, relative angle, derived from normal vectors, exhibits properties similar to normal vectors, both of which demonstrate stability in undamaged sections and high randomness in damaged sections. To evaluate the extent to which the relative angle retains the characteristics of normal vectors, and to compare the distribution of different features across different sections, this study employs an entropy-based feature evaluation. It quantifies the amount of information or the uncertainty associated with random variables and has been used to evaluate and filter complex features in many studies (Huo et al. 2020; Odhiambo Omuya et al. 2021).

For continuous random variables, their entropy values are calculated as:

$$\mathbf{H}(X) = -\int_{-\infty}^{\infty} f(x) \log f(x)\, dx \tag{15}$$

, where $f(x)$ is the probability density function (PDF) of $x$. In this study, the point cloud datasets contain the position coordinates, the normal vectors, and the relative angles of the points. They are stored in the datasets as one or more scalars within a certain range.

To simplify the calculation, this study discretises the data, breaking the continuous range of values into several bins. The probabilities of data falling within each bin are calculated by counting the number of points in each bin and dividing by the total number of points. The entropy calculation is modified as:

$$\mathbf{H}(X) = -\sum_{i=1}^{n} p_i \log p_i \tag{16}$$

, where $n$ is the number of total bins, $p_i$ is the probability associated with the $i$-th bin.

Among the three features of points, the position coordinates and normal vectors each contain three independent variables, while the relative angles have only one variable.

## 3 Experimental setup

### 3.1 Specimen and data collection

The concrete surface damage data in this study was collected from 22 standard concrete cylindrical specimens ($\Phi 100 \times 200$ mm$^2$). All specimens were fabricated following a standard process and cured for 28 days before being destroyed under compression tests. The defects on these specimens include cracks and spalling, which are rich in morphological variety. Cylindrical specimens were chosen because generating complex damage features on them is economical and convenient. In subsequent data processing, the curved cylinder surface will be unfolded to simulate a flat concrete surface.

The surface damage of the specimens was captured using a handheld 3D scanner (Artec Space Spider with 3D accuracy of 0.05 mm and 3D resolution of 0.1 mm), which reconstructs the target objects based on multi-camera stereo vision reconstruction. The scanning was performed indoors under stable lighting conditions to ensure consistent data quality. During the process, the specimen was slowly rotated on a turntable to ensure that the captured point cloud was uniformly and stably distributed. Simultaneously, the operator held the scanner, scanned and then reconstructed the side surface of the specimens where the defects were present. The entire scanning procedure was repeated using the same steps for each specimen, ensuring consistency in data collection. Fig. 3 shows the scanner and the scanning operation.

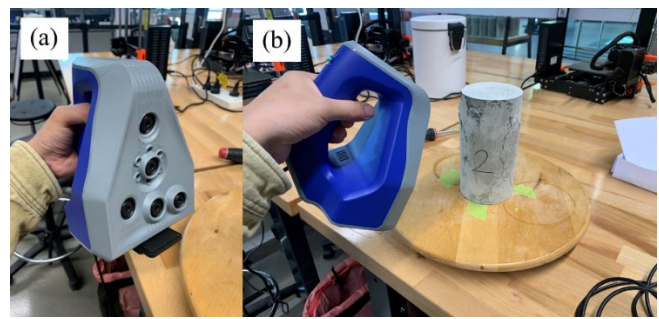

Fig. 3 (a) handheld 3D scanner, (b) scanning operation

## 3.2 Data preprocessing and point cloud division

In this study, to better approximate the flat surfaces typically found in general concrete structures, the 3D reconstructions of concrete cylinders are unfolded into planar representations. Data annotation is consequently based on unfolded point clouds and divided into two categories: damaged and undamaged sections.

Generally, to fit the acquired point clouds into a neural network model, it is necessary to divide the obtained point clouds into small subsets with a fixed number of points. In this study, the total number of points, $N_{all}$, in the unfolded point cloud is obtained. Assuming the neural network requires an input with $n_{input}$ points, the number of subsets, $num_s$, that a complete point cloud should be divided into can be calculated as follows:

$$num_s = \left\lfloor \frac{N_{num}}{n_{input}} \right\rfloor \tag{17}$$

Then, farthest point sampling (FPS) (Eldar et al. 1997) is applied on the unfolded point clouds to find $num_s$ points as reference points for subsets of point clouds. For each reference point, $n-1$ nearest neighbouring points are found using the KD tree method instead of using the distance from points to the reference points, and these $n$ points are extracted as a subset of the complete reconstruction.

Fig. 4 presents the cross-sectional profile of a typical crack and illustrates the differences in sampling results that may arise from using different point selection methods. For the reference point shown in Fig. 4(a), the distance-based point selection calculates the absolute distances from all points to this reference point and selects them sequentially based on their distances. In this case, the selected points may not be directly connected to the reference point, as shown in Fig. 4(b). On the one hand, the normal vectors of the isolated parts may not be correctly obtained, which affects the calculation of the relative angle. On the other hand, because the isolated parts are disconnected from the whole, their relative positions in the sample are inconsistent with those of other regular points. It can be considered that the features of the isolated parts, no matter whether the isolated parts are annotated as damaged or not, differ from those of the vast majority of other parts of the point clouds. To avoid the impact of inconsistencies on training and incorrect feature calculation, the KD tree method is used to obtain neighbouring points of the reference point, as shown in Fig. 4(c), which can avoid isolated parts in sampling.

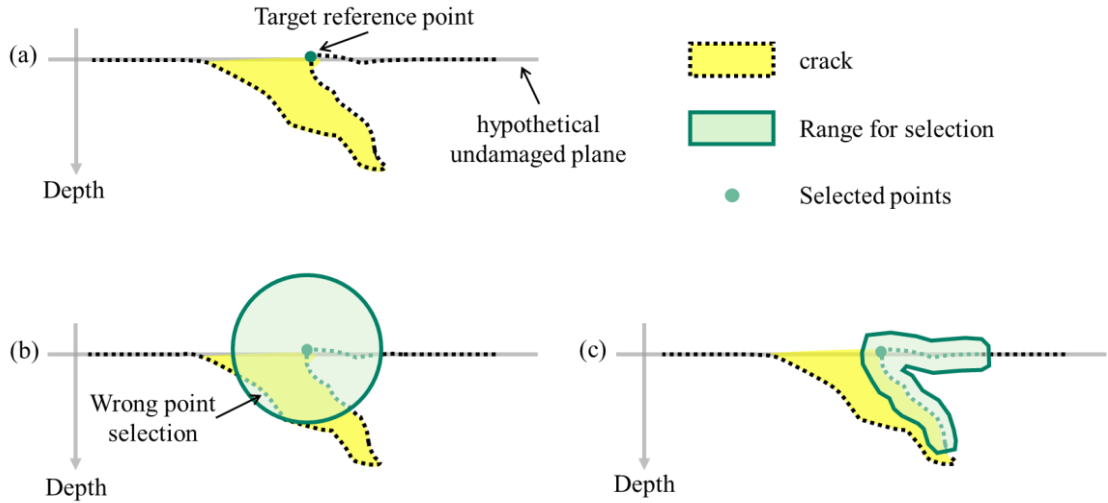

Fig. 4 (a) cross-section, (b) point selection based on distance, (c) point selection using the KD tree method

The three-dimensional information in the reconstructions obtained through scanning includes the position coordinates of points. To examine the effectiveness of the proposed feature, the divided subsets will be processed according to section 2 to obtain the position coordinates, normal vectors, and relative angles of the points under different normalisation methods.

### 3.3 Feature evaluation and model performance evaluation

This study compares the entropy of position coordinates, normal vectors, and relative angles of all collected point clouds in the damaged and undamaged sections. It is implemented according to the entropy-based feature evaluation described in section 2.5. This evaluation focuses on the uncertainty in information of different features in different regions of the samples. The two normalisation methods described in section **Error! Reference source not found.** will affect the calculation of normal vectors. Therefore, point clouds obtained based on different normalisation methods will also be evaluated separately.

Besides, this study will evaluate the performance of different features and feature combinations in PointNet++ (Qi et al. 2017). PointNet++ is a popular neural network designed for point cloud recognition and segmentation tasks, utilising max pooling layers to enable the model to effectively extract information from unordered 3D point clouds. It can capture differences among local regions without requiring an alignment pre-processing step, yet still achieve strong performance. PointNet++ is widely adopted as a baseline for comparison in 3D segmentation applications, delivering results close to state-of-the-art models in various applications (Zhang et al. 2023). In this study, since the primary focus is on how different features and feature combinations influence model performance, only PointNet++ is employed for training and comparison.

In this study, the hyperparameters were empirically chosen as shown in Table 1.

Table 1 Hyperparameters and searching range

| Config. | Value |
|---|---|
| Optimiser | Adam |
| criteria | Negative log likelihood |
| Learning rate (lr) | 1e-3 |
| Weight decay | 1e-4 |
| Decay step for lr decay | 10 |
| Decay rate for lr decay | 0.7 |
| Batch size | 32 |
| epochs | 32 |

This study applied three quantitative evaluation indicators, Accuracy, Mean Intersection over Union (mIoU), and IoUs for damaged and undamaged sections, to assess the performance of PointNet++ using different features and feature combinations.

Accuracy measures the proportion of correctly predicted points out of the total number of points. It is defined as:

$$\text{Accuracy} = \frac{TP + TN}{TP + TN + FP + FN} \tag{18}$$

, where TP, TN, FP, and FN are true positive, true negative, false positive, and false negative samples, respectively. In this study, positive samples refer to damaged sections and false samples refer to undamaged sections.

mIoU is the average intersection over union across all classes. IoU for a single class is defined as the ratio of the intersection of the predicted and ground truth sets to their union.

$$IoU_c = \frac{\text{Intersection Area}}{\text{Union Area}} = \frac{|P_c \cap G_c|}{|P_c \cup G_c|} \tag{19}$$

$$mIoU = \frac{1}{C}\sum_{c=1}^{C} IoU_c \tag{20}$$

, where $P_c$ refers to the point sets predicted as class $c$, $G_c$ refers to the ground truth point sets of class $c$, and $C$, the total number of classes, is two for damaged and undamaged sections in this study.

The model training was coded in Python and executed on a desktop with an Intel Core i7-12700F 2.1 GHz processor, NVIDIA GeForce RTX 3080 10 G graphics card, and 32 GB RAM.

## 4    Results and discussion

This section shows the results and discussion on specimen reconstruction and evaluation of feature combinations. Key comparisons of storage efficiency and model performance using various feature combinations are reported.

## 4.1 Reconstruction and subdivision of specimens

Through the preprocessing described in section 3.2, 22 complete reconstructions of the cylinder side surfaces were obtained. Considering the efficiency and feasibility of model training, the inputs to the model were the subsets of complete reconstructions composed of 4096 points, and the division of the reconstructions was then determined according to section 3.2. The number of points in a single specimen reconstruction ranges from 474,830 to 475,000. Thus, one reconstruction is divided into around 114 subsets, and there are 2530 subsets in total for 22 specimens. Based on this division, all subsets contain 10,362,880 points in total, with 2,943,091 points in damaged sections and 7,419,789 points in undamaged sections. An overview of the dataset parameters and divisions is provided in Table 2. Fig. 5 shows several typical unfolded reconstructions and their annotations. It is noted that cracks and spalling are distributed throughout reconstructions. Fig. 6 shows the reference points of a typical reconstruction, the location of one particular reference point, and the range of its corresponding subset. There is one subset presented in Fig. 6(d). It can be observed that this subset consists of the reference point and the nearby connected points. Since the unfolded reconstruction is approximately planar, the subset appears nearly circular when viewed head-on. However, due to the unevenness of the reconstruction surface, the subset does not present as a perfect circle.

Table 2 Dataset configuration and division

| Config. | Value |
| --- | --- |
| Number of samples | 22 |
| Points in each sample | 474,830 to 475,000 |
| Total points | 10,362,880 |
| Points in each network input | 4096 |
| Number of subsets | 2530 |
| Damaged:undamaged | 2,943,091: 7,419,789 ($\approx$ 28.4%: 71.6%) |
| Train:validation:test | 8:1:1 |

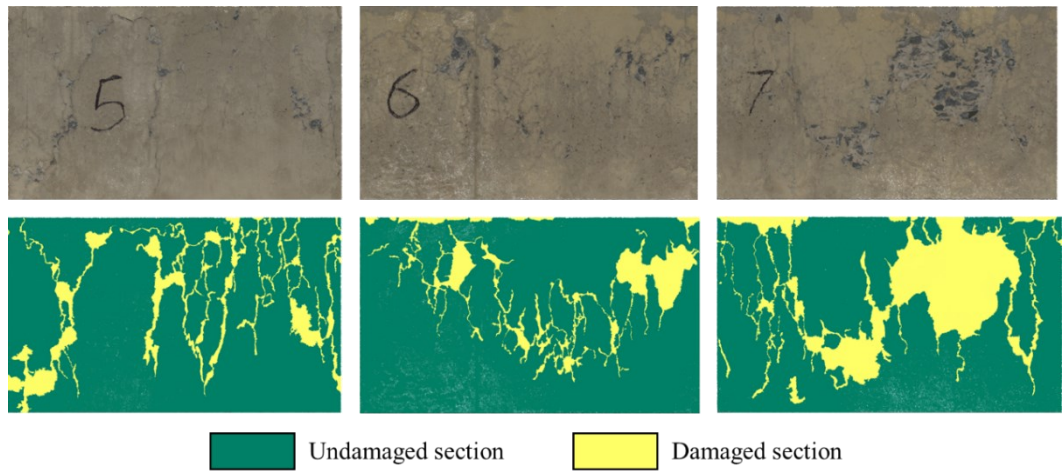

Fig. 5 Unfolded cylinder reconstructions and annotations

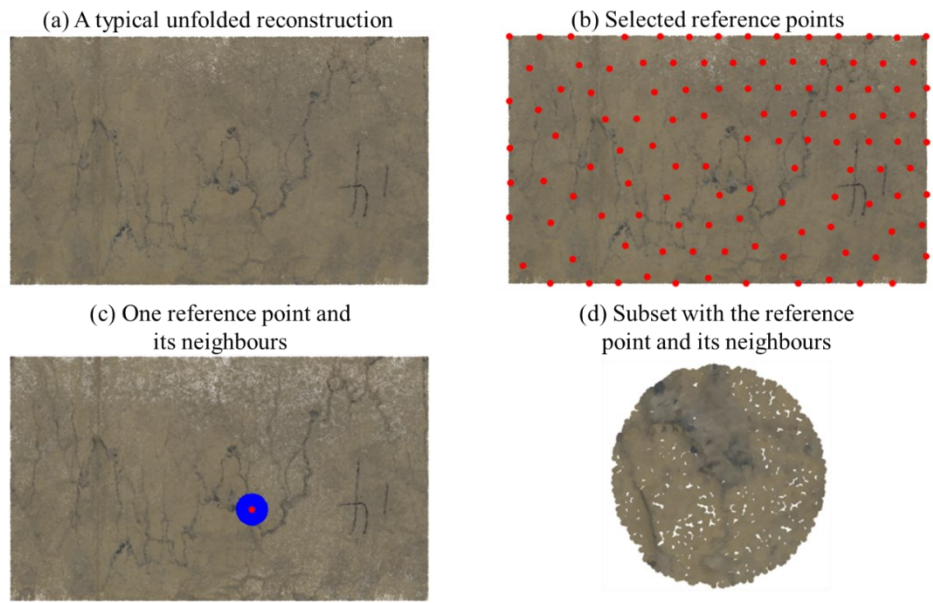

Fig. 6 (a) a typical unfolded cylinder reconstruction, (b) selected reference points of one reconstruction (114 subsets in total), (c) one of the reference points and the range of its neighbours, (d) the subset of the reference point shown in (c)

A subset of cylinder point clouds, processed by the feature calculations described in section 2, has three features: position coordinate, normal vector, and relative angle. These three features can be obtained from point clouds after global and axis-specific normalisations respectively. Therefore, for one subset of point clouds, two sets of features can be acquired under different normalisation operations. In addition, in point cloud data, the position coordinates which contain the spatial information of points are the most widely used feature and the basis for extracting other features in most related applications. Therefore, in this study, the position coordinates of the points will be included in all of the feature combinations. Notably, even if the position coordinates are not directly used as training features, they must be retained in all feature combinations for mapping purposes.

## 4.2 Data storage and feature combinations

Table 3 lists the feature combinations employed in this study and the corresponding average required storage space for one subset data file with annotations. Each subset data file serves as a single training or inference sample. It should be noted that, in Combinations 5 and 6, the position coordinates are used exclusively for mapping purposes. Position coordinates in these two combinations aim to indicate the spatial locations of points, but are not included as input features to the network for computation. Among all these feature combinations, "position coordinates + normal vector" (Combination 2) have been reported in many studies to achieve high performance (Daghigh et al. 2022; Mirzaei et al. 2022; Xu and Stilla 2021). Therefore, it is the 'base' combination for comparing the required space as shown in Table 3.

As the proposed feature, "relative angle", contains only one dimension, and the normal vector has three dimensions, Combinations 3 and 5 reduce the storage requirement by 27.6% compared to Combination 2. This makes the proposed feature, relative angle, more efficient for use in practical scenarios where hardware conditions are usually strictly limited. In this research, with a total amount of 2530 subsets reconstructed from 22 specimens, using the widely used feature combination of position coordinates and normal vector required a total of 662 MB of storage space. In contrast, when using position coordinates and relative angle, only 427 MB of space was needed, alleviating the potential issue of insufficient storage space in practice.

Besides, the proposed relative angle feature offers a crucial additional advantage that the input dimensionality can be significantly reduced. Traditional methods utlising position coordinates combined with three-dimensional normal vectors (Combination 2) require six input feature channels, three for position coordinates and three for vector components. In contrast, Combination 3 operates with only four input channels. This represents a 33% reduction in input dimensionality compared to the conventional baseline. More significantly, when using the position representation only for mapping purposes alongside the relative angle (Combination 5), the input complexity is reduced to a single feature channel. This combination achieves an 83% reduction in required input channels compared to the six-channel base case.

Table 3 Feature combinations and required storage space for one reconstruction subset

| Feature combination | Storage requirement (kB) | % to base in storage | Number of input feature channels | % to base in input feature channels |
|---|---|---|---|---|
| 1. Position coordinate | 156 | 58.2% | 3 | 50% |
| 2. Position coordinate + Normal vector (base) | 268 | 100% | 6 | 100% |
| 3. Position coordinate + Relative angle | 194 | 72.4% | 4 | 66.7% |
| 4. Position coordinate + Normal vector + Relative angle | 306 | 114.2% | 7 | 116.7% |
| 5. Position coordinate (for mapping) + Relative angle | 194 | 72.4% | 1 | 16.7% |
| 6. Position coordinate (for mapping) + Normal vector | 268 | 100% | 3 | 100% |

## 4.3 Feature entropy

The number of bins of feature evaluation was set to 10 for all extracted features: normalised position coordinates, normal vectors, and relative angles. For each reconstruction subset, the entropies of features were estimated, and then the average value of all subsets was taken to estimate the entropies across all point clouds.

Table 4 shows the entropies under two normalisation operations. As position coordinates and normal vectors are features composed of three variables each, the corresponding blocks display the entropies for each of these three variables separately. The average of these three entropy values is shown in parentheses, representing the overall entropy of that feature across all subsets.

Table 4 Entropies of features for undamaged and damaged sections under different normalisation

| | | Overall | Undamaged section | Damaged section |
|---|---|---|---|---|
| Global normalisation $(x, y, z)/s_g$ | Position coordinate (XYZ) | 3.22, 3.24, 3.04 (3.17) | 3.24, 3.25, 3.06 (3.18) | 3.02, 3.15, 3.01 (3.06) |
| | Normal vector | 2.39, 1.95, 1.76 (2.03) | 1.99, 1.49, 1.25 (1.58) | 3.14, 2.92, 2.71 (2.92) |
| | Relative angle | 1.98 | 1.43 | 3.05 |
| Axis-specific normalisation $\left(\frac{x}{s_x}, \frac{y}{s_y}, \frac{z}{s_z}\right)$ | Position coordinate (XYZ) | 3.22, 3.24, 3.04 (2.74) | 3.24, 3.25, 3.06 (3.18) | 3.02, 3.15, 3.01 (3.06) |
| | Normal vector | 2.98, 2.69, 2.84 (2.84) | 2.81, 2.41, 2.51 (2.58) | 3.19, 3.21, 3.21 (3.20) |
| | Relative angle | 2.75 | 2.46 | 3.09 |

The entropy-based feature evaluation results show that the position coordinates have high entropy under both normalisation methods, demonstrating the richness of information in it. Additionally, the difference in entropy of position coordinates between damaged and undamaged parts is insignificant, which may indicate that the diversity contained in the point positions is similar across different sections. It suggests that using position coordinates as a feature, in terms of numerical distribution, does not show distinct characteristics between damaged and undamaged areas. This is because the distribution of coordinates within their respective ranges does not change, regardless of the normalisation method.

Normal vectors and relative angles both exhibit lower entropy in undamaged sections and higher entropy in damaged sections. It might indicate that both features can simplify information in undamaged sections while preserving information in damaged sections. Moreover, it is worth noting that the relative angle feature has a similar entropy to the normal vector across all settings. In undamaged parts, the angle feature demonstrates even lower entropy. This suggests that the relative angle has a comparable ability to the normal vector in simplifying information in undamaged sections.

Additionally, it is noted that both normal vectors and relative angles demonstrate more effective simplification of undamaged sections under global normalisation. It can be seen that under axis-specific normalisation, although the normal vectors and angles still show a significant simplification of information in undamaged sections with lower entropy than in damaged sections, it is much higher than the corresponding values under global normalisation. There are two possible reasons for this:

One possible reason is that under axis-specific normalisation, small shape changes in the undamaged sections are amplified, leading to more significant variations in the orientations of normal vectors, thus resulting in relatively higher entropy in the observed sections.

Another possible reason is that the density of points in specific directions may have changed under axis-specific normalisation. It leads to differences in the selection of neighbouring points when calculating normal vectors compared to global normalisation, potentially magnifying the angular changes in normal vectors.

### 4.4 Model training and performance under different feature combinations

The subsets of reconstructions (2530 in total) were randomly assigned into training, test, and evaluation datasets with an 8:1:1 ratio for the training of PointNet++. When subsets of reconstructions were fed into the model training as inputs, they underwent a random rotation to ensure that the network performance was not limited to the directions of the collected data, ensuring generalisation ability. The network will be trained under six different feature combinations as shown in Table 5. These combinations include the use of position coordinates together with both the relative angle and normal vector, as well as training results obtained by using each feature independently. For the convenience of illustration, in Table 5, 'XYZ' represents the position coordinates, 'Normal' represents the normal vector, and 'Angle' represents the relative angle. In addition, the subscript of 'axis' represents that the feature is derived from the point cloud under axis-specific normalisation, and the subscript of 'global' indicates that the feature is derived from the point cloud processed by global normalisation.

Table 5 Model performance using different feature combinations

| No. | Feature combination | Accuracy | mIoU | IoU of undamaged section | IoU of damaged section |
|---|---|---|---|---|---|
| 1-1 | $XYZ_{axis}$ | 0.908 | 0.811 | 0.900 | 0.741 |
| 1-2 | $XYZ_{global}$ | 0.920 | 0.833 | 0.902 | 0.768 |
| 2-1 | $XYZ_{axis}+Normal_{axis}$ | 0.920 | 0.824 | 0.907 | 0.783 |
| 2-2 | $XYZ_{axis}+Normal_{global}$ | 0.929 | 0.858 | 0.916 | 0.811 |
| 2-3 | $XYZ_{global}+Normal_{axis}$ | 0.923 | 0.830 | 0.905 | 0.773 |
| 2-4 | $XYZ_{global}+Normal_{global}$ | 0.944 | 0.871 | 0.926 | 0.822 |
| 3-1 | $XYZ_{axis}+Angle_{axis}$ | 0.928 | 0.851 | 0.907 | 0.791 |
| 3-2 | $XYZ_{axis}+Angle_{global}$ | 0.932 | 0.848 | 0.909 | 0.794 |
| 3-3 | $XYZ_{global}+Angle_{axis}$ | 0.931 | 0.847 | 0.916 | 0.784 |
| 3-4 | $XYZ_{global}+Angle_{global}$ | 0.942 | 0.870 | 0.920 | 0.822 |
| 4-1 | $XYZ_{axis}+Normal_{axis}+Angle_{axis}$ | 0.945 | 0.866 | 0.929 | 0.804 |
| 4-2 | $XYZ_{global}+Normal_{global}+Angle_{global}$ | 0.936 | 0.864 | 0.920 | 0.824 |
| 5-1 | $Angle_{axis}$ (with $XYZ_{axis}$) | 0.924 | 0.830 | 0.899 | 0.761 |
| 5-2 | $Angle_{global}$ (with $XYZ_{global}$) | 0.947 | 0.884 | 0.926 | 0.843 |
| 6-1 | $Normal_{axis}$ (with $XYZ_{axis}$) | 0.922 | 0.819 | 0.899 | 0.740 |
| 6-2 | $Normal_{global}$ (with $XYZ_{global}$) | 0.949 | 0.889 | 0.930 | 0.849 |

Table 5 groups the same feature combinations together to compare their performance under different normalisation methods and highlights in bold the best-performing feature combination within each group. For the first case, using only position coordinates, i.e., 'XYZ', the performance under global normalisation, which does not change the original shape of the subsets, is better than the one provided under axis-specific normalisation. The difference between these two is negligible in the undamaged section, while there is a 2.7% IoU difference in the damaged section. This may suggest that using the original geometric shape of the target might be more advantageous in concrete damage identification. The same trend can be observed in the results of all the combinations.

'XYZ+Normal' and 'XYZ+Angle' use the normal vector or relative angle to enhance the recognition of targets. It can be observed that both feature combinations can effectively improve the model performance, and both achieve optimal results under global normalisation (Combinations 2-4 and 3-4). It is worth noting that 'XYZ+Angle' and 'XYZ+Normal' can achieve similar performance, but according to Table 3, 'XYZ+Angle' only requires 72.4% of the storage space of the latter. Compared to those using only position coordinates, Combinations 2-4 and 3-4 increased the overall accuracy by 2.4% and 2.2%, respectively. Notably, they show more improvement in recognising the damaged section, with a 5.4% increase in IoU, while for the undamaged section, their IoU only increased by 2.4% and 1.8%. This may be partially because the IoU for the undamaged section was around 90%, in which case, further improvement might be challenging.

On the other hand, according to the feature evaluation results in Table 4, both normal vector and relative angle have significantly different entropy values in damaged and undamaged sections compared to position coordinates. It indicates that the variability of information in the two sections differs noticeably, thus having the potential to support more effective feature extraction and analysis. It is observed that although Combinations 2-2 and 3-2 use $Normal_{global}$ and $Angle_{global}$, they do not lead to an improvement as high as Combinations 2-4 and 3-4 can achieve. This might suggest that using $XYZ_{axis}$ impacts the effective utilisation of the other two features. Besides, $Normal_{axis}$ and $Angle_{axis}$ show improvements in model performance, but less than those obtained under global normalisation. This issue matches the results of the feature entropy evaluation, where normal vector and relative angle under axis-specific normalisation, although showing low entropy in the undamaged section, have a more minor difference in entropy values between undamaged and damaged sections than those under global normalisation.

The fourth feature combination in Table 5 includes all features under global normalisation. Compared to Combinations 2-4 and 3-4, Combination 4 uses more features but does not show significant differences in model performance. This indicates that in the scenario involved in this study, adding normal vectors and relative angles to position coordinates of points may not be necessary.

Combinations 5 and 6 show the training results using only the relative angle or the normal vector as features. In these combinations, position coordinates, labelled as "$XYZ_{axis}$" or "$XYZ_{global}$" are excluded from training and serve only as spatial references for mapping outputs. As shown, the single-dimensional relative angle achieves performance comparable to the three-dimensional normal vector. More notably, when position coordinates are not included in the feature set, both the relative angle and the normal vector demonstrate better overall performance. Their overall accuracies are close to those of Combinations 2–4 and 3–4, but they improve the IoU of damaged sections by more than 2%. This indicates that both the relative angle and normal vector offer a more effective representation of concrete damage than position coordinates, consistent with the observations in Table 4. Furthermore, the relative angle is particularly attractive, delivering similar performance with enhanced efficiency due to its lower dimensionality. As shown in Table 3, Combination 5 only uses one input channel, which is 16.7 % compared to 6-channel baselines. Under identical network architectures and without point cloud resampling, this compact representation directly decreases arithmetic intensity requirements and alleviates memory bandwidth pressure during inference, allowing concurrent processing of larger point cloud batches within fixed hardware constraints.

Additionally, the performance of Combinations 5.1 and 6.1 is significantly worse than that of Combinations 5.2 and 6.2, and even inferior to Combinations 2.1 and 3.1 which use the same data composition but include position coordinates in training. This suggests that when training relies only on relative angle or normal vector features, preserving the original geometric morphology of concrete damaged surfaces and conducting feature selection and analysis based on such a geometry becomes crucial. The geometric changes induced by axis-specific normalisation seem to have a significantly negative impact on training outcomes.

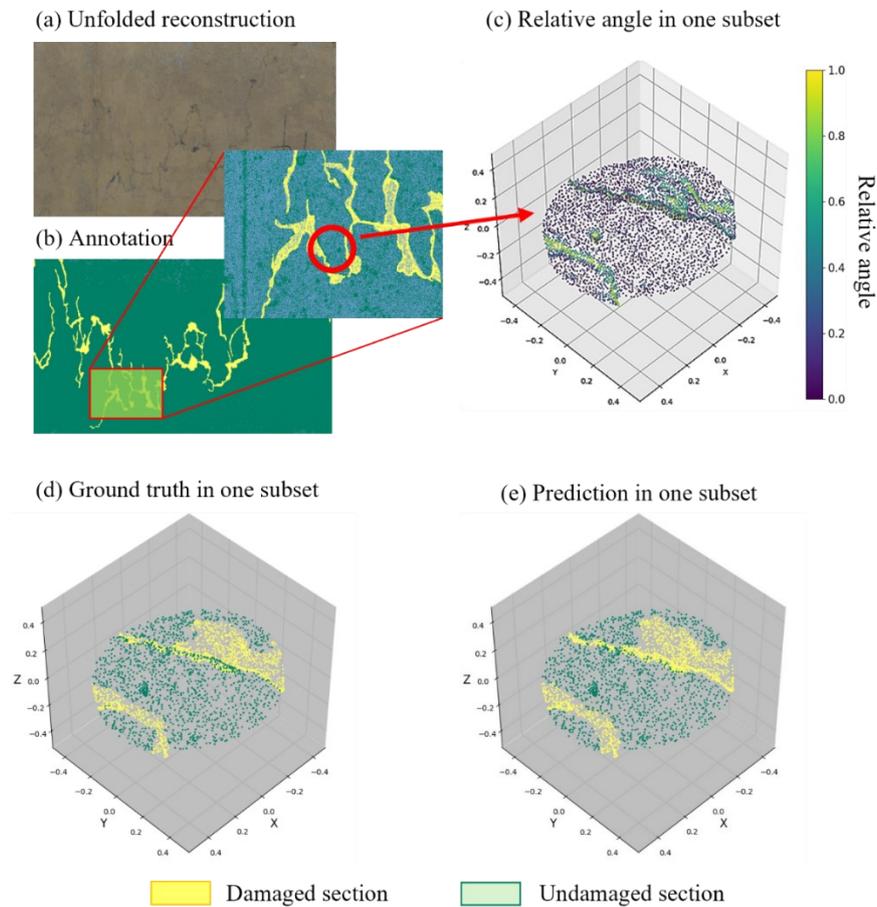

Fig. 7 (a) a typical unfolded reconstruction, (b) annotation and zoom-in view for one subset, (c) the subset showing relative angle values, (d) ground truth for the subset, (e) prediction for the subset

Fig. 7 shows a typical unfolded reconstruction of a cylinder specimen and one of its subsets. Fig. 7(c) displays the relative angle values of the points in that subset. It can be seen that the relative angle has uniform values in undamaged regions, but shows some variation in damaged sections, which matches the ranges of labelled damaged and undamaged sections in Fig. 7(d). Fig. 7(e) shows the prediction results for that subset using the model trained with Combination 5, 'Angle$_{global}$'. It can be observed that the predicted results closely match the labelled results, with only some misclassifications at the edges of the damage.

## 5    Conclusion

This study proposes a new feature, relative angle, for identifying damage on concrete surfaces. This feature is applicable for point cloud-based feature learning and representation, calculated based on normal vectors of points. Compared with combining position coordinates and normal vectors as features, using relative angles to replace normal vectors only requires 72.4% of the storage space and maintains comparable model performance. It greatly reduces data storage and corresponding computational resources for concrete defect inspection applications, contributing to miniaturisation and cost reduction for structural inspection devices. This study offers several significant contributions as follows:

1. A new feature called 'relative angle' is proposed. This feature is derived by computing the angle between the normal vector of a point and the average normal vector of its parent point cloud. It provides directionality information equivalent to three-dimensional normal vectors for concrete surface defect characteristics. When deployed with compact position mapping where position coordinates serve solely for spatial mapping, the proposed feature reduces 27.6% of the storage space and the total input channels per point from six to one (an 83% reduction), outperforming the conventional feature combination with position coordinate and normal vector. This low-dimensional representation achieves storage footprint reduction across all processing stages and enables larger-batch execution on resource-constrained devices using identical network architectures.

2. An entropy-based feature evaluation is employed to assess different features, examining the uncertainty and degree of variation in information. The results demonstrate that both relative angle and normal vector exhibit differences in entropy between damaged and undamaged sections, with these differences becoming more significant under global normalisation. In contrast, position coordinates, serving as the basic feature of points in the point cloud, do not manifest such differences.

3. The study combines position coordinates, normal vectors, and relative angles obtained under different normalisation methods for PointNet++ training. In this widely used model for point cloud data, relative angle demonstrates performance comparable to normal vector, especially under global normalisation. This indicates that relative angle as a feature can practically improve model performance, with similar functionality and effectiveness to normal vectors. As it requires lower hardware demands compared to normal vectors, it provides a new option for the data structures in the practical deployment of algorithms in designing portable devices for concrete defect detection.

However, the proposed feature was only tested in highly controlled settings, which leads to certain limitations in this study, motivating the following future work:

1. This study aims to compare the impact of different features and feature combinations on model performance. Consequently, one representative model, PointNet++, was selected for training under various feature combinations. Further validation across multiple relevant models is necessary for future work. Moreover, the data in this study was collected from concrete cylinder specimens whose damage characteristics may differ from actual structures. Future work will extend testing to different base surface geometries, roughness levels, or material colours and patterns, to ensure broader applicability of the findings in diverse practical scenarios.

2. Due to limitations of the division method, although each subset divided from the point cloud is based on different reference points with certain distances from others, there may still be slight overlaps at the edges of the subsets. This can limit the generalisation ability of the trained model in real-world scenarios. Moreover, the scale and precision of the point cloud data used in this study may not fully capture the complexity of larger or lower-resolution datasets, raising questions about the reliability under varying resolutions and broader scenarios. Future research will focus on collecting data from diverse scenarios, refining the data division strategy to mitigate overlaps, and systematically evaluating the model performance across different scales and resolutions.

**Data availability statement**

Some or all data, models, or code generated or used during thestudy are proprietary or confidential in nature and may only be pro-vided with restrictions.